\documentclass[sigconf]{acmart}
\AtBeginDocument{%
  }

\setcopyright{acmlicensed}
\copyrightyear{2018}
\acmYear{2018}
\acmDOI{XXXXXXX.XXXXXXX}
\acmConference[Conference acronym 'XX]{Make sure to enter the correct
  conference title from your rights confirmation email}{June 03--05,
  2018}{Woodstock, NY}
\acmISBN{978-1-4503-XXXX-X/2018/06}

\acmSubmissionID{}

\begin{document}

\title{THETA: Heat Diffusion as a Structural Prior for Cross-Domain Transfer in Graph Foundation Models}

\author{Haonan Liu}
\authornote{Both authors contributed equally to this research.}
\affiliation{%
  \institution{University of Electronic Science and Technology of China}
  \city{Chengdu}
  \country{China}}
\email{liuhaonan025@gmail.com}

\author{Shuo Wang}
\authornotemark[1]
\affiliation{%
  \institution{Tsinghua University}
  \city{Beijing}
  \country{China}}
\email{runner21st@gmail.com}

\author{Zhao Kang}
\correspondingauthor
\affiliation{%
  \institution{University of Electronic Science and Technology of China}
  \city{Chengdu}
  \country{China}}
\email{zkang@uestc.edu.cn}

\renewcommand{\shortauthors}{Trovato et al.}

\begin{abstract}
Graph foundation models (GFMs) aim to learn transferable representations across domains with heterogeneous node attributes and graph structures. However, relations inferred from node attributes can rely on domain-specific feature semantics, limiting their transferability to unseen graphs, particularly in low-label regimes. We propose \textbf{THETA} (\textbf{T}opology-aware \textbf{HE}at-based \textbf{T}r\textbf{A}nsfer), a graph foundation model that leverages \textbf{Heat Kernel Signatures (HKS)} as a topology-based structural prior for cross-domain transfer. By characterizing nodes' heat-return behavior across diffusion scales, HKS provides a common structural descriptor across graphs without requiring aligned attribute semantics, complementing feature-based relations with information about local and multiscale connectivity. THETA incorporates this prior into both pre-training and few-shot adaptation. During pre-training, it combines HKS with node attributes to construct refined graph views and trains a shared encoder via contrastive learning. During adaptation, it uses HKS to guide edge-weight adjustment and prototype refinement while keeping the encoder frozen and retaining the contribution of labeled support nodes. Through HKS, THETA uses structural information from heat diffusion to guide both representation learning and few-shot adaptation with unlabeled target-domain data. Extensive experiments across diverse graph domains demonstrate that THETA consistently outperforms existing baselines on downstream tasks, highlighting the value of heat diffusion as a structural prior for transferable graph representations.

\end{abstract}



\received{20 February 2007}
\received[revised]{12 March 2009}
\received[accepted]{5 June 2009}

\maketitle

\section{Introduction}

Graphs support many Web applications, representing hyperlinks between pages, interactions among users, and associations among products~\cite{brin1998web, ying2018pinsage}. These graphs differ in their node attributes, topology, and prediction tasks~\cite{wu2021gnnsurvey}. Training a separate model for each domain requires repeated effort and task-specific annotations, which may be scarce in a new domain~\cite{pan2010transfer}. Graph foundation models (GFMs) seek to address this challenge by learning knowledge that transfers across graph domains and tasks~\cite{yu2025samgpt,shi2024gfm}.

Recent work pursues transfer through several complementary strategies. UniGraph uses language-based representations to accommodate heterogeneous graph attributes~\cite{unigraph2025}. GCOPE connects source graphs through learnable coordinator nodes, while SAMGPT introduces structure tokens and prompts for multi-domain pre-training and target adaptation~\cite{zhao2024gcope,yu2025samgpt}. These approaches broaden the range of graphs that a shared model can process. However, language-based interfaces require usable textual descriptions~\cite{unigraph2025}, and aligning feature dimensions alone cannot establish shared attribute meanings~\cite{pan2010transfer,scgfm2026}.

Structural modeling offers another basis for cross-domain learning~\cite{franceschi2019lds,scgfm2026}. MDGFM combines feature and topology information to refine graph relations~\cite{MDGFM}. RiemannGFM learns structural knowledge through tree and cycle vocabularies in Riemannian spaces, while R-GFM combines multiscale subgraphs with Riemannian Graph-of-Graphs modeling~\cite{sun2025riemanngfm,liu2026rgfm}. SCGFM uses learnable geometric bases and Gromov--Wasserstein alignment to construct shared structural coordinates~\cite{scgfm2026}. These advances enrich structural representations and cross-graph alignment. They also motivate a further question: how can information computed directly from graph topology provide explicit guidance for both relation learning and adaptation to an unseen domain?

This challenge arises at two stages of cross-domain learning: constructing relations for message passing~\cite{kipf2017gcn,franceschi2019lds} and selecting unlabeled target nodes to refine class prototypes~\cite{snell2017prototypical,ren2018semisupervised}. Both decisions require evidence about which nodes are useful to one another. When these decisions rely on attributes or learned representation similarities, they can remain sensitive to domain-specific feature semantics~\cite{dai2022graphtransfer}. For example, MDGFM estimates structural relations using neighborhood-aggregated attributes, so this source of structural evidence still depends on the input features~\cite{MDGFM}. Structural vocabularies, geometric alignment, and multiscale encoding enrich representations, but do not by themselves ensure that the resulting similarities identify useful relations for a new task. Under few-shot scenarios, there is limited supervision to correct unsuitable propagation relations or assess retrieved unlabeled examples~\cite{ren2018semisupervised}. This motivates an additional structural reference that can be computed without aligned attribute semantics and remain available throughout pre-training and adaptation.

Heat diffusion provides such a reference. The heat response at a node over time reflects return patterns induced by the surrounding connectivity~\cite{donnat2018graphwave}. \textbf{H}eat \textbf{K}ernel \textbf{S}ignatures (HKS) summarize this behavior across diffusion times, providing multiscale node descriptions computed solely from topology~\cite{sun2009concise}. Their common form allows structural comparisons without requiring attributes to have the same meaning across domains~\cite{donnat2018graphwave,tsitsulin2018netlsd}. To account for differences in graph spectra, we select graph-specific diffusion times using shared normalized heat-trace targets, providing a common reference for scale selection. Exact HKS descriptors are equivariant to node relabeling and admit perturbation bounds at fixed diffusion times. These properties motivate its use as an explicit \emph{structural prior}: it supplies evidence about connectivity that complements feature-based relations. Structural similarity does not necessarily imply a shared class, so we combine this prior with node attributes, learned representations, and labeled support examples.

Building on this idea, we propose \textbf{THETA} (\textbf{T}opology-aware \textbf{HE}at-based \textbf{T}r\textbf{A}nsfer), a graph foundation model that uses HKS as a topology-derived structural prior. During pre-training, HKS and node attributes jointly guide the construction of refined graph views, from which a shared encoder learns through contrastive learning~\cite{you2020graphcl}. During few-shot adaptation, the encoder remains frozen, while HKS guides edge-weight adjustment and complements embedding similarity when retrieving unlabeled nodes for prototype refinement. Labeled support nodes remain explicit contributors to the class prototypes. By using the same topology-derived prior across both stages, THETA connects relation construction during pre-training with propagation adjustment and support-based prediction in unseen domains.

Our main contributions are summarized as follows:
\begin{itemize}
    \item We propose THETA, a cross-domain graph foundation
    model built around heat diffusion as an explicit
    structural prior. Multiscale HKS provides topology-based
    node descriptions that complement attributes without
    requiring shared feature semantics.

    \item We incorporate this prior into both multi-domain
    pre-training and few-shot adaptation. HKS guides
    graph-view construction upstream and edge-weight
    adjustment and unlabeled-node retrieval downstream,
    connecting relation learning with target-specific
    adaptation.

    \item We evaluate THETA on cross-domain few-shot node
    classification and supervised link prediction.
    Ablations examine the prior's contribution in both
    learning stages, while further analyses characterize
    cross-scale HKS variation and the balance between
    graph propagation and node attributes.
\end{itemize}

\section{Related Work}
\label{sec:related_work}

\subsection{Graph Foundation Models}

Existing GFMs support cross-domain transfer through shared attribute interfaces, multi-domain pre-training, and lightweight adaptation. UniGraph uses language-based representations to accommodate heterogeneous graph attributes \cite{unigraph2025}. GCOPE connects source graphs through learnable coordinator nodes \cite{zhao2024gcope}, while SAMGPT uses structure tokens and prompts for multi-domain learning and target adaptation \cite{yu2025samgpt}.

Another line of work emphasizes transferable structural knowledge. MDGFM, which directly informs our learning framework, combines graph structure learning with multi-domain pre-training and prompt-based adaptation \cite{MDGFM}. Its relation estimation uses attributes and neighborhood-aggregated features. THETA builds on this framework by introducing an additional structural reference computed independently of node attributes. Other methods organize structural knowledge through geometric representations: RiemannGFM models tree and cycle structures in Riemannian spaces \cite{sun2025riemanngfm}; SCGFM learns geometric bases and aligns graphs through Gromov--Wasserstein distances \cite{scgfm2026}; and R-GFM constructs a multi-scale Graph-of-Graphs from subgraphs at different hop distances \cite{liu2026rgfm}. These methods highlight the importance of structural alignment and scale in cross-domain learning, motivating our investigation of heat diffusion as a source of explicit structural guidance.

\subsection{Heat Diffusion and Structural Descriptors}

Heat diffusion captures graph connectivity at multiple scales.
Originally developed for shape analysis, HKS summarizes
heat-return behavior across diffusion times~\cite{sun2009concise}.
GraphWave~\cite{graphwave} characterizes node structural roles
through distributions of heat diffusion wavelet coefficients,
while NetLSD~\cite{tsitsulin2018netlsd} constructs graph-level signatures
from heat-kernel traces.
These methods demonstrate how diffusion yields structural
descriptors independently of node attributes.
Building on this foundation, THETA calibrates diffusion
times for each graph and incorporates the resulting HKS
into cross-domain learning as an explicit structural prior.
This prior guides relation learning during pre-training,
as well as edge adaptation and prototype refinement
in the target domain.

\section{Preliminaries}
\label{sec:preliminaries}

\begin{figure*}[t]
  \centering
  \includegraphics[width=\textwidth]{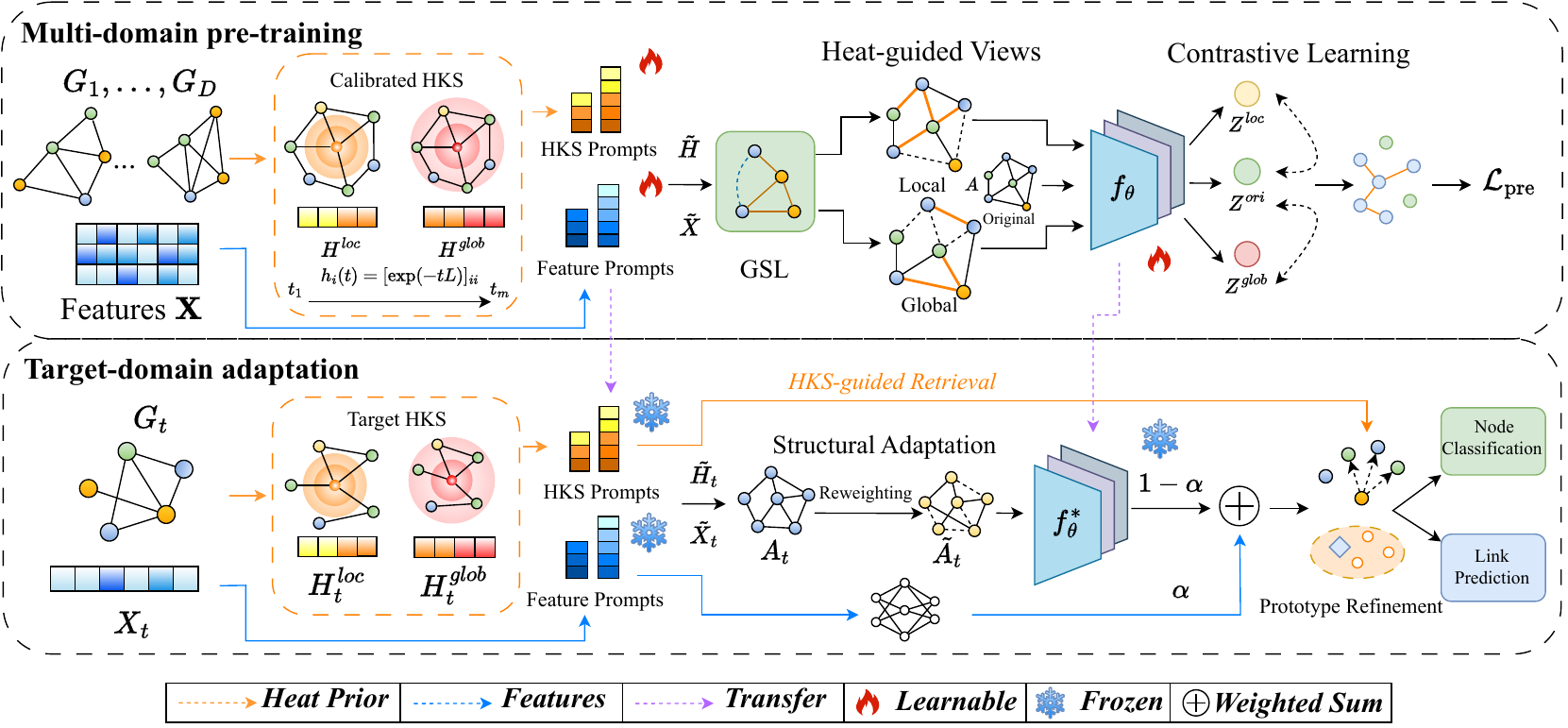}
  \caption{The overall framework of the proposed THETA}
  \label{pipeline}
\end{figure*}

\subsection{Problem Formulation}
\label{sec:problem_setup}

Let $G_d=(V_d,E_d,\mathbf{X}_d)$ be a graph from domain $d$,
with node set $V_d$, edge set $E_d$, node attributes
$\mathbf{X}_d\in\mathbb{R}^{n_d\times p_d}$, and adjacency
$\mathbf{A}_d\in\mathbb{R}^{n_d\times n_d}$, where $n_d=|V_d|$.
Attribute dimensions, feature semantics, and label spaces may
differ across domains.

Given source graphs
$\mathcal{G}_{\mathrm{src}}=\{G_d\mid d\in\mathcal{D}_{\mathrm{src}}\}$,
we pre-train a shared encoder $f_\theta$ without node labels,
then adapt the model with target-domain supervision for
few-shot node classification or link prediction on an unseen
graph $G_t$ from domain $d_t\notin\mathcal{D}_{\mathrm{src}}$.
Notation is summarized in Appendix~\ref{app:notation}.

\subsection{Heat Diffusion and Heat Kernel Signatures}
\label{sec:heat_diffusion}

For clarity, we omit the domain subscript and write $G=(V,E,\mathbf{X})$ with $n=|V|$. For heat diffusion, we use a symmetric, nonnegative adjacency matrix $\mathbf{A}$, obtained by symmetrizing the observed adjacency when necessary. Let
$\mathbf{D}$ be the diagonal degree matrix, with
$D_{vv}=\sum_u A_{vu}$. The symmetric normalized graph Laplacian is
\begin{equation}
    \mathbf{L}
    =
    \mathbf{I}
    -
    \mathbf{D}^{-1/2}\mathbf{A}\mathbf{D}^{-1/2},
    \label{eq:normalized_laplacian}
\end{equation}
where $D_{vv}^{-1/2}=0$ for zero-degree nodes.

Heat diffusion follows
$\partial\mathbf{z}(t)/\partial t=-\mathbf{L}\mathbf{z}(t)$,
where $\mathbf{z}(t)$ denotes the heat signal at time $t$.
Its solution is
$\mathbf{z}(t)=\mathbf{K}_t\mathbf{z}(0)$, with the heat kernel
\begin{equation}
    \mathbf{K}_t
    =
    \exp(-t\mathbf{L})
    =
    \sum_{i=1}^{n}
    e^{-t\lambda_i}\mathbf{u}_i\mathbf{u}_i^\top,
    \label{eq:heat_kernel}
\end{equation}
where $(\lambda_i,\mathbf{u}_i)$ are the eigenpairs of
$\mathbf{L}$ and the eigenvectors are orthonormal.

The Heat Kernel Signature (HKS) characterizes a node through the diagonal entries of the heat kernel~\cite{sun2009concise}. At diffusion time $t$, the signature of node $v$ is
\begin{equation}
    h_v(t)
    =
    [\mathbf{K}_t]_{vv}
    =
    \sum_{i=1}^{n}
    e^{-t\lambda_i}u_i(v)^2,
    \label{eq:hks_definition}
\end{equation}
where $u_i(v)$ is the component of $\mathbf{u}_i$ at node $v$.
Starting from a unit heat impulse at $v$, this quantity measures the heat response at the same node after time $t$.

The structural meaning of HKS can also be seen through a walk expansion. Let $\mathbf{S}=\mathbf{D}^{-1/2}\mathbf{A}\mathbf{D}^{-1/2}$, so that $\mathbf{L}=\mathbf{I}-\mathbf{S}$. Then
\begin{equation}
    h_v(t)
    = e^{-t}\sum_{k=0}^{\infty}
    \frac{t^k}{k!}[\mathbf{S}^k]_{vv}.
    \label{eq:hks_walk_expansion}
\end{equation}
Here, $[\mathbf{S}^k]_{vv}$ sums the weights of length-$k$ walks that start and end at $v$, with each step weighted by the corresponding entry of $\mathbf{S}$. HKS therefore summarizes return patterns determined by the surrounding connectivity. The diffusion time controls the weights assigned to different walk lengths, providing structural information that does not depend on node attributes.

Given an ordered sequence of diffusion times
$\mathcal{T}=(t_1,\ldots,t_M)$ with $0\leq t_1\leq\cdots\leq t_M$, we form the multiscale descriptor
\begin{equation}
    \mathbf{h}_v
    =
    [h_v(t_1),\ldots,h_v(t_M)]^\top
    \in\mathbb{R}^{M}.
    \label{eq:multiscale_hks}
\end{equation}
Stacking these descriptors gives
$\mathbf{H}\in\mathbb{R}^{n\times M}$.
As $t$ increases, spectral components with larger eigenvalues decay more rapidly. Evaluating HKS at multiple times therefore captures node connectivity through different spectral weightings.

HKS depends only on graph topology and is equivariant to node relabeling: at the same diffusion times, a permutation of the nodes produces the same permutation of the rows of $\mathbf{H}$. It thus provides a structural description without requiring aligned attribute semantics across domains. THETA uses this description as a prior alongside node attributes and task-specific supervision. Section~\ref{sec:structural_prior} describes how we select diffusion times for each graph and construct the structural inputs used by the model.

\section{Methodology}

\subsection{Overview of THETA}
\label{sec:overview}

Figure~\ref{pipeline} illustrates the overall framework of THETA.
We use multiscale HKS computed from graph topology as an
explicit structural prior for multi-domain pre-training and
target-domain adaptation. During pre-training, HKS complements
node attributes in constructing refined graph views, from which
a shared encoder learns through contrastive learning.
During few-shot adaptation, the encoder remains frozen, while
HKS guides edge-weight adjustment and the retrieval of unlabeled
nodes for prototype refinement. For link prediction, HKS
supplements node representations used for edge scoring.
The following sections detail prior construction, pre-training,
and target-domain adaptation.

\subsection{Constructing the Heat-Diffusion Structural Prior}
\label{sec:structural_prior}

THETA constructs a structural prior from each graph's observed topology. Since the same diffusion time can correspond to different stages on graphs with different spectra, we select graph-specific times using shared heat-trace targets. The resulting HKS descriptors provide structural inputs at shorter and longer diffusion scales.

\textbf{Selecting diffusion scales.}
Let $m$ be the multiplicity of the zero eigenvalue of the symmetric normalized Laplacian $\mathbf{L}$. For a graph with $n>m$ nodes, we define
\begin{equation}
    r_G(t)
    =\frac{\operatorname{tr}(\exp(-t\mathbf{L}))-m}{n-m}.
    \label{eq:normalized_heat_trace}
\end{equation}
Subtracting $m$ removes the stationary contribution, while dividing by $n-m$ averages over the decaying spectral modes. Thus, $r_G(t)$ decreases from $1$ toward $0$ and provides a graph-level measure of diffusion progress.

We choose $M\geq2$ geometrically spaced targets between shared bounds $0<\rho_{\min}<\rho_{\max}<1$, where the upper and lower bounds correspond to earlier and later diffusion stages, respectively:
\begin{align}
    \rho_j
    &=\rho_{\max}
    \left(\frac{\rho_{\min}}{\rho_{\max}}\right)^{\frac{j-1}{M-1}},
    \quad j=1,\ldots,M,
    \label{eq:heat_trace_targets}\\
    t_{G,j}
    &=\min\left\{r_G^{-1}(\rho_j),\,t_{\max}\right\}.
    \label{eq:calibrated_diffusion_times}
\end{align}
Here, $t_{\max}>0$ caps diffusion times, yielding a nondecreasing sequence $\mathcal{T}_G=(t_{G,1},\ldots,t_{G,M})$. For uncapped coordinates, graphs share the same normalized heat-trace target but may use different numerical times. This supplies a common reference for scale selection without assuming node correspondences across domains. Truncation and degenerate cases are detailed in Appendix~\ref{app:hks_calibration}.

\textbf{Forming multiscale node descriptors.}
Using $h_v(t)$ from Eq.~\eqref{eq:hks_definition}, we divide the ordered scales into two groups:
\begin{equation}
    \begin{aligned}
        \mathbf{h}_v^{(s)}
        &=[h_v(t_{G,1}),\ldots,h_v(t_{G,M_s})]^\top,\\
        \mathbf{h}_v^{(l)}
        &=[h_v(t_{G,M_s+1}),\ldots,h_v(t_{G,M})]^\top,
    \end{aligned}
    \label{eq:two_scale_hks}
\end{equation}
where $M_s$ and $M_l=M-M_s$ are the numbers of shorter- and longer-time coordinates, with $1\leq M_s<M$. Longer times place greater relative weight on slowly decaying spectral modes, allowing the two groups to describe connectivity through different spectral weightings. Stacking over nodes gives $\mathbf{H}^{(s)}\in\mathbb{R}^{n\times M_s}$ and $\mathbf{H}^{(l)}\in\mathbb{R}^{n\times M_l}$.

These priors require neither attributes nor labels and remain fixed during learning. Their prompt-modulated forms, $\widetilde{\mathbf{H}}^{(s)}$ and $\widetilde{\mathbf{H}}^{(l)}$, adapt how structural coordinates contribute to relation estimation. We use eigendecomposition for small graphs and polynomial approximation with stochastic estimation for larger graphs; numerical details are given in Appendix~\ref{app:hks_numerical}.

\textbf{Stability of the structural prior.}
The following property bounds the sensitivity of exact HKS descriptors to changes in graph structure at fixed diffusion times.

\textbf{Proposition (Fixed-time stability).}
Let $\mathbf{L}$ and $\widetilde{\mathbf{L}}$ be symmetric positive semidefinite graph Laplacians on the same ordered node set, with $\|\mathbf{L}-\widetilde{\mathbf{L}}\|_2\leq\epsilon$. For any common nonnegative time sequence $\mathcal{T}=(t_1,\ldots,t_M)$,
\begin{equation}
    \|\mathbf{h}_v(\mathbf{L};\mathcal{T})
    -\mathbf{h}_v(\widetilde{\mathbf{L}};\mathcal{T})\|_2
    \leq\epsilon\left(\sum_{j=1}^{M}t_j^2\right)^{1/2},
    \label{eq:hks_fixed_time_stability}
\end{equation}
where matrix and vector norms are spectral and Euclidean norms, respectively.

At fixed scales, small Laplacian perturbations therefore produce bounded changes in the structural descriptors, supporting their use as a stable reference for relation learning. Graph-specific time recalibration introduces an additional variation term. The proof, the recalibration analysis, and equivariance under node relabeling are provided in Appendix~\ref{app:hks_properties}.

\subsection{HKS-Guided Multi-Domain Pre-training}
\label{sec:pretraining}

THETA combines the structural prior from Section~\ref{sec:structural_prior} with node attributes to learn node relations. THETA uses it to construct two graph views, allowing structural similarity to influence which nodes exchange information during pre-training. A shared encoder then learns representations from the original graph and these refined views.

\textbf{Preparing attribute and structural inputs.}
We first project each domain's node attributes to a common dimension using PCA. For each branch $b\in\{s,l\}$, separate prompts modulate the projected attributes and the corresponding HKS descriptors. Let $\mathbf{q}$ denote either a node's PCA-projected attribute vector or its HKS descriptor in branch $b$. Omitting branch and input-type indices, the prompt takes the form
\begin{equation}
    \mathcal{P}_d(\mathbf{q})
    =\omega\mathbf{q}
    +(1-\omega)(\mathbf{p}_{\mathrm{sh}}\odot
    \mathbf{p}_d\odot\mathbf{q}),
    \label{eq:source_prompt}
\end{equation}
where $\mathbf{p}_{\mathrm{sh}}$ is shared across source domains, $\mathbf{p}_d$ is domain-specific, $\odot$ denotes elementwise multiplication, and $\omega\in(0,1)$ is a learned scalar parameterized by a sigmoid. The prompt tokens have the same dimension as the corresponding input $\mathbf{q}$. Each branch and input type has its own prompt parameters. We denote the resulting attribute and structural matrices by $\widetilde{\mathbf{X}}^{(b)}$ and $\widetilde{\mathbf{H}}^{(b)}$, respectively. PCA provides dimensional compatibility, while the prompts allow the model to adjust the contribution of individual attribute and diffusion coordinates.

\textbf{Learning relations with HKS.}
For each source graph, we construct a candidate set
$\mathcal{E}_{\mathrm{cand}}^{(b)}=E\cup\mathcal{E}_x^{(b)}\cup\mathcal{E}_h^{(b)}$,
where $\mathcal{E}_x^{(b)}$ and $\mathcal{E}_h^{(b)}$ contain nearest-neighbor relations under attribute cosine similarity and HKS Gaussian similarity, respectively. Including HKS neighbors allows nodes with similar heat-return patterns to become candidates even when they are not close in attribute space. For each candidate pair $(u,v)$, we compute
\begin{equation}
    \begin{aligned}
        s_{uv}^{x,b}
        &=\cos(\widetilde{\mathbf{x}}_u^{(b)},
        \widetilde{\mathbf{x}}_v^{(b)}),\\
        s_{uv}^{h,b}
        &=\exp\left(-\frac{
        \|\widetilde{\mathbf{h}}_u^{(b)}
        -\widetilde{\mathbf{h}}_v^{(b)}\|_2^2}
        {\tau_h}\right),
    \end{aligned}
    \label{eq:attribute_hks_relations}
\end{equation}
where $\tau_h>0$ controls the HKS similarity scale. A gate
$g_{uv}^{(b)}=\sigma(\operatorname{MLP}_{\psi_b}(\mathbf{c}_{uv}^{(b)}))$
combines the two signals. Its input $\mathbf{c}_{uv}^{(b)}$ contains the attributes and HKS of both endpoints, their mean neighborhood descriptors, and normalized node degrees. Neighborhood descriptors are computed from the observed graph. The fused relation and its residual combination with the observed edge weight are
\begin{equation}
    \begin{aligned}
        s_{uv}^{(b)}
        &=g_{uv}^{(b)}s_{uv}^{x,b}
        +(1-g_{uv}^{(b)})s_{uv}^{h,b},\\
        w_{uv}^{(b)}
        &=(1-\zeta)A_{uv}+\zeta s_{uv}^{(b)},
    \end{aligned}
    \label{eq:hks_relation_fusion}
\end{equation}
where $\zeta\in(0,1)$ controls the contribution of learned relations and $A_{uv}=0$ for candidate pairs outside the observed edge set. This makes the influence of HKS depend on the node pair and its context, while retaining a contribution from the observed topology.

\textbf{Structural views and contrastive learning.}
We retain the top-$k$ candidate weights for each node, add reverse edges, and average reciprocal weights. Self-loops and degree normalization produce the refined adjacency $\mathbf{A}^{(b)}$; edge dropout is applied during training. The shorter- and longer-time HKS branches give two refined views. Writing $\mathbf{A}^{(o)}=\operatorname{Norm}(\mathbf{A})$ for the original adjacency with self-loops and degree normalization, the shared encoder produces
\begin{equation}
    \begin{aligned}
        \mathbf{Z}^{(o)}
        &=f_{\theta}(\widetilde{\mathbf{X}}^{(s)},\mathbf{A}^{(o)}),\\
        \mathbf{Z}^{(b)}
        &=f_{\theta}(\widetilde{\mathbf{X}}^{(b)},\mathbf{A}^{(b)}),
        \quad b\in\{s,l\}.
    \end{aligned}
    \label{eq:pretraining_views}
\end{equation}
HKS thus enters pre-training through relation construction, while the encoder processes the prompted node attributes on each view.

For two distinct views $\nu,\nu'\in\{o,s,l\}$, the same node forms a positive pair and other nodes in the comparison set $\mathcal{B}$ form negative pairs. Let $c_{vu}^{\nu,\nu'}=\cos(\mathbf{z}_v^{(\nu)},\mathbf{z}_u^{(\nu')})$. The directional contrastive loss is
\begin{equation}
    \ell_{\nu\rightarrow\nu'}
    =-\frac{1}{|\mathcal{B}|}
    \sum_{v\in\mathcal{B}}
    \log\frac{\exp(c_{vv}^{\nu,\nu'}/\tau_{\mathrm{ctr}})}
    {\sum_{u\in\mathcal{B}}\exp(c_{vu}^{\nu,\nu'}/\tau_{\mathrm{ctr}})},
    \label{eq:directional_contrastive_loss}
\end{equation}
where $\tau_{\mathrm{ctr}}>0$ is the contrastive temperature. We use the symmetric loss
\begin{equation}
    \mathcal{L}_{\mathrm{ctr}}
    (\mathbf{Z}^{(\nu)},\mathbf{Z}^{(\nu')})
    =\frac{\ell_{\nu\rightarrow\nu'}
    +\ell_{\nu'\rightarrow\nu}}{2}.
\end{equation}
The pre-training objective is
\begin{equation}
    \mathcal{L}_{\mathrm{pre}}
    =\sum_{d\in\mathcal{D}_{\mathrm{src}}}
    \sum_{b\in\{s,l\}}\lambda_b
    \mathcal{L}_{\mathrm{ctr}}
    (\mathbf{Z}_d^{(o)},\mathbf{Z}_d^{(b)}),
    \label{eq:pretraining_objective}
\end{equation}
with nonnegative weights $\lambda_b$. Contrastive pairs are formed within each source graph. The encoder, branch-specific relation estimators, and shared prompt tokens are trained jointly across domains, while domain-specific tokens provide separate modulation. Through this objective, HKS-guided relations shape the views from which the shared encoder learns.

\subsection{HKS-Guided Target-Domain Adaptation}
\label{sec:target_adaptation}

On an unseen target graph, THETA computes HKS from the observed topology and learns target-specific prompts while keeping the source prompt tokens fixed. These prompts combine source knowledge with target-specific modulation of attributes and HKS (Appendix~\ref{app:target_prompts}). We average the prompted attribute branches to obtain $\overline{\mathbf{x}}_v$ and concatenate the prompted HKS branches to obtain $\overline{\mathbf{h}}_v$. The underlying HKS descriptors remain fixed.

\textbf{HKS-guided structural adaptation.}
For few-shot node classification, we freeze the pre-trained encoder and adapt its propagation using the structural prior. For each observed edge $(u,v)$, a relation vector $\mathbf{d}_{uv}$ summarizes attribute and HKS similarities between the endpoints and their neighborhoods. A lightweight modifier adjusts the edge weight:
\begin{equation}
    \begin{aligned}
        \delta_{uv}
        &=\tanh(\operatorname{MLP}_{\psi_t}(\mathbf{d}_{uv})),\\
        A^{\mathrm{adapt}}_{uv}
        &=A_{uv}(1+\delta_{uv}).
    \end{aligned}
    \label{eq:target_edge_adjustment}
\end{equation}
Attribute-nearest-neighbor edges supplement the observed topology. The frozen GCN processes the adapted graph, while a trainable MLP independently processes the prompted attributes. Their outputs are combined as
\begin{equation}
    \mathbf{z}_v
    =(1-\alpha)\mathbf{z}_v^{\mathrm{GCN}}
    +\alpha\mathbf{z}_v^{\mathrm{MLP}},
    \qquad \alpha\in[0,1].
    \label{eq:target_node_embeddings}
\end{equation}
Here, $\alpha$ controls the balance between graph propagation and node attributes. HKS guides propagation through edge-weight adjustment, while the MLP provides a complementary representation without neighborhood aggregation.

\textbf{HKS-guided prototype refinement.}
Let $\mathcal{S}_c$ denote the labeled support nodes of class $c$ and $\mathcal{U}$ the unlabeled target nodes. To enrich support-based prototypes, we retrieve unlabeled nodes using both heat-diffusion structure and learned representations. For a support node $s$ and candidate $u\in\mathcal{U}$, the retrieval score is
\begin{equation}
    \begin{aligned}
        s_{us}^{\mathrm{aug}}
        &=\lambda_{\mathrm{aug}}
        \cos(\overline{\mathbf{h}}_u,\overline{\mathbf{h}}_s)\\
        &\quad+(1-\lambda_{\mathrm{aug}})
        \cos(\mathbf{z}_u,\mathbf{z}_s),
    \end{aligned}
    \label{eq:prototype_retrieval_score}
\end{equation}
where $\lambda_{\mathrm{aug}}\in[0,1]$. HKS thus complements representation similarity with evidence about surrounding connectivity.

For each support node, we select up to $N$ highest-scoring candidates. Let $\mathcal{R}_c$ collect the resulting support--candidate pairs $(s,u)$ for class $c$, with weights
$w_{us}=\exp(s_{us}^{\mathrm{aug}}/\tau_{\mathrm{aug}})$.
The refined prototype is
\begin{equation}
    \mathbf{p}_c
    =
    \frac{
    w_c^{\mathrm{sup}}\sum_{s\in\mathcal{S}_c}\mathbf{z}_s
    +\sum_{(s,u)\in\mathcal{R}_c}w_{us}\mathbf{z}_u
    }{
    w_c^{\mathrm{sup}}|\mathcal{S}_c|
    +\sum_{(s,u)\in\mathcal{R}_c}w_{us}
    },
    \label{eq:hks_refined_prototype}
\end{equation}
where $w_c^{\mathrm{sup}}=2\max_{(s,u)\in\mathcal{R}_c}w_{us}$ preserves each support node's contribution with a larger individual weight than any retrieved candidate. Retrieval and weighting use only HKS descriptors and node representations, without accessing candidate labels. Only labeled support nodes provide class supervision. When refinement is disabled or no candidates are available, we use the support-only mean prototype.

\textbf{Task objectives and prediction.}
For node classification, we optimize the target prompts, edge modifier, and MLP using cross-entropy over support-node cosine scores $\cos(\mathbf{z}_v,\mathbf{p}_c)$, with the encoder frozen. Query nodes are assigned to the most similar prototype. For link prediction, a task-specific projection combines node attributes, encoder representations, and HKS descriptors; cosine similarity between projected endpoints provides the edge score. We train with binary cross-entropy on positive and negative node pairs, updating the encoder, target prompts, and projection while keeping source prompt tokens fixed.

\section{Experiments}
\label{sec:experiments}

\begin{table*}[t]
    \centering
    \renewcommand{\arraystretch}{0.88}
    \setlength{\aboverulesep}{1pt}
    \setlength{\belowrulesep}{1pt}
    \caption{
        Accuracy (\%) on cross-domain 1-shot node classification, reported as mean $\pm$ standard deviation. The best results are highlighted in bold, and the runner-up with an underline.
    }
    \label{tab:node_classification_1shot}

    \resizebox{\textwidth}{!}{%
    \begin{tabular}{@{}lcccccccc@{}}
        \toprule
        Method
        & Wisconsin & Cornell & CiteSeer & Cora
        & PubMed & Computers & Photo & Texas \\
        \midrule

        \multicolumn{9}{l}{\textit{GNN Baselines}} \\
        \midrule
        GCN
        & $17.46 \pm 3.35$ & $19.53 \pm 5.77$
        & $26.89 \pm 3.01$ & $31.98 \pm 4.11$
        & $44.29 \pm 2.22$ & $39.43 \pm 5.68$
        & $50.39 \pm 1.19$ & $18.48 \pm 4.08$ \\

        GAT
        & $16.86 \pm 3.16$ & $16.51 \pm 8.41$
        & $25.27 \pm 2.10$ & $26.81 \pm 4.42$
        & $45.11 \pm 5.52$ & $38.05 \pm 4.47$
        & $56.51 \pm 2.35$ & $18.36 \pm 4.63$ \\

        \midrule
        \multicolumn{9}{l}{
            \textit{Self-Supervised Pre-training with Fine-tuning}
        } \\
        \midrule
        DGI
        & $28.24 \pm 5.30$ & $29.73 \pm 10.11$
        & $34.52 \pm 8.81$ & $39.88 \pm 5.53$
        & $47.10 \pm 4.95$ & $34.59 \pm 5.94$
        & $50.66 \pm 6.93$ & $29.73 \pm 6.34$ \\

        GraphCL
        & $32.16 \pm 8.83$ & $31.35 \pm 12.47$
        & $27.56 \pm 5.01$ & $35.26 \pm 8.62$
        & $43.66 \pm 3.12$ & $34.40 \pm 6.93$
        & $48.86 \pm 5.53$ & $31.35 \pm 22.66$ \\

        \midrule
        \multicolumn{9}{l}{\textit{Prompt-based Adaptation}} \\
        \midrule
        GPPT
        & $23.53 \pm 15.93$ & $25.41 \pm 5.27$
        & $33.24 \pm 8.73$ & $40.60 \pm 6.00$
        & $41.64 \pm 9.29$ & $44.57 \pm 13.38$
        & $53.94 \pm 11.93$ & $18.38 \pm 12.00$ \\

        GPF
        & $23.53 \pm 9.36$ & $36.22 \pm 9.46$
        & $27.16 \pm 4.10$ & $34.74 \pm 3.48$
        & $26.62 \pm 10.18$ & $42.15 \pm 8.75$
        & $59.13 \pm 6.05$ & $28.65 \pm 17.31$ \\

        GraphPrompt
        & $20.39 \pm 7.92$ & $28.65 \pm 5.92$
        & $19.80 \pm 2.79$ & $33.68 \pm 9.83$
        & $42.94 \pm 1.98$ & $33.34 \pm 6.09$
        & $42.60 \pm 6.75$ & $24.32 \pm 9.17$ \\

        \midrule
        \multicolumn{9}{l}{\textit{Graph Foundation Models}} \\
        \midrule
        GCOPE
        & $28.69 \pm 7.00$ & $25.00 \pm 6.58$
        & $38.62 \pm 2.58$ & $39.05 \pm 2.53$
        & $44.52 \pm 2.51$ & $28.50 \pm 4.91$
        & $59.17 \pm 2.45$ & $19.50 \pm 12.18$ \\

        SAMGPT
        & $21.57 \pm 12.09$ & $36.22 \pm 5.30$
        & $28.78 \pm 8.99$ & $43.08 \pm 10.40$
        & $37.95 \pm 9.58$ & $45.80 \pm 4.54$
        & $49.80 \pm 11.89$ & $31.89 \pm 11.26$ \\

        MDGFM
        & $17.65 \pm 7.34$ & $26.49 \pm 10.73$
        & $23.32 \pm 15.32$ & $42.88 \pm 9.12$
        & $38.42 \pm 7.92$ & $42.68 \pm 2.53$
        & $57.78 \pm 9.07$ & $31.54 \pm 11.78$ \\

        RiemannGFM
        & $31.20 \pm 9.35$ & $31.35 \pm 7.76$
        & $31.86 \pm 6.45$ & $43.48 \pm 5.28$
        & $44.10 \pm 5.05$ & $46.20 \pm 8.79$
        & $57.20 \pm 16.84$ & $29.12 \pm 7.44$ \\

        R-GFM
        & $\underline{35.41 \pm 7.29}$ & $\underline{36.71 \pm 9.92}$
        & $\boldsymbol{57.54 \pm 9.49}$ & $\underline{49.50 \pm 3.97}$
        & $\underline{49.80 \pm 5.38}$ & $\underline{52.30 \pm 3.33}$
        & $\underline{61.08 \pm 5.26}$ & $\underline{32.36 \pm 12.10}$ \\

        \midrule
        \textbf{THETA}
        & $\boldsymbol{42.18 \pm 9.91}$ & $\boldsymbol{40.26 \pm 7.29}$ 
        & $\underline{45.00 \pm 9.67}$ & $\boldsymbol{52.22 \pm 7.56}$ 
        & $\boldsymbol{53.18 \pm 7.52}$ & $\boldsymbol{53.10 \pm 8.05}$ 
        & $\boldsymbol{62.45 \pm 8.73}$ & $\boldsymbol{33.84 \pm 6.03}$ \\
        \bottomrule
    \end{tabular}%
    }
\end{table*}

We evaluate THETA on unseen graph domains, focusing on both
downstream performance and the role of heat diffusion as a
structural prior. Our experiments address five research questions:
\textbf{RQ1:} How does THETA compare with existing methods on
cross-domain few-shot node classification?
\textbf{RQ2:} Can THETA transfer to supervised link prediction
on unseen target graphs?
\textbf{RQ3:} How do graph structure learning and the heat-based
prior contribute to pre-training and target-domain adaptation?
\textbf{RQ4:} How do redundancy and variation across calibrated
HKS scales differ between graphs?
\textbf{RQ5:} How does the balance between graph propagation
and feature-based adaptation affect performance across target domains?

\subsection{Experimental Setup}
\label{sec:experimental_setup}

\textbf{Datasets.}
We use ten graphs from four domains: citation networks
(Cora, CiteSeer~\citep{sen2008collective}, and PubMed~\citep{namata2012query}), Webpage networks
(Cornell, Texas, and Wisconsin)~\citep{CravenFMMNS98}, Wikipedia networks
(Chameleon and Squirrel)~\citep{rozemberczki2021multi,pei2020geom}, and product co-purchase networks
(Computers and Photo)~\citep{shchur2018pitfalls}. Following the benchmark of R-GFM~\citep{liu2026rgfm},
we adopt leave-one-dataset-out transfer, pre-training on the
remaining graphs without node labels and evaluating on the
held-out target. Dataset statistics are
provided in Appendix~\ref{app:datasets}.

\textbf{Baselines.}
We compare THETA with four groups of methods:
supervised GNNs (GCN~\citep{kipf2017semi} and GAT~\citep{velivckovic2018graph}), self-supervised methods
(DGI~\citep{velickovic2019dgi} and GraphCL~\citep{you2020graphcl}), prompt-based methods
(GPPT~\citep{sun2022gppt}, GPF~\citep{fang2023universal}, and GraphPrompt~\citep{GraphPrompt}), and GFMs
(GCOPE~\citep{zhao2024gcope}, SAMGPT~\citep{yu2025samgpt}, MDGFM~\citep{MDGFM}, RiemannGFM~\citep{sun2025riemanngfm}, and R-GFM~\citep{liu2026rgfm}).
Brief descriptions of these groups are provided
in Appendix~\ref{app:baselines}.

\textbf{Tasks and metrics.}
Our main task is cross-domain few-shot node classification
with $K\in\{1,3,5\}$ labels per class on eight target graphs.
We additionally evaluate supervised link prediction on
Wisconsin, Cora, PubMed, and Photo.
We report mean accuracy for node classification and mean
ROC-AUC for link prediction, together with standard deviations.
Evaluation protocols are detailed in
Appendix~\ref{app:evaluation_protocols}.

\textbf{Implementation.}
THETA uses a three-layer GCN, PCA-based attribute alignment, and
16 HKS coordinates covering short and long diffusion scales.
The encoder is frozen during few-shot node classification
and updated during supervised link prediction.
Hyperparameters and training details are provided in
Appendix~\ref{app:implementation}. Our code is available at \url{https://anonymous.4open.science/r/THETA}.

\subsection{Few-Shot Node Classification (RQ1)}
\label{sec:fewshot_results}

Tables~\ref{tab:node_classification_1shot}
and~\ref{tab:node_classification_5shot} summarize the results.
THETA achieves the highest mean accuracy on seven of eight
target datasets in both settings, with gains across citation,
Webpage, and product co-purchase networks.
Compared with R-GFM, it improves accuracy on Wisconsin by
6.77 and 9.89 percentage points under 1-shot and 5-shot
evaluation, respectively, and on Computers by 7.87 points
under 5-shot evaluation.
The strong Webpage results are notable given their
heterophilous connectivity, suggesting that THETA's overall
adaptation procedure can accommodate targets where neighboring
nodes frequently belong to different classes.

Increasing the support budget from one to five labels per
class improves THETA's accuracy on all eight targets,
demonstrating its ability to benefit from additional
target-domain supervision.
The advantages persist under 5-shot evaluation, where
prototypes use only labeled support nodes, indicating that
the gains extend beyond the 1-shot prototype-refinement
configuration.
CiteSeer remains an exception, with THETA ranking second
behind R-GFM, suggesting that transfer effectiveness remains
target-dependent.
Overall, these results support THETA's cross-domain
effectiveness; the contribution of heat as a structural
prior is examined separately in
Section~\ref{sec:rq3}.

\begin{table*}[t]
    \centering
    \renewcommand{\arraystretch}{0.88}
    \setlength{\aboverulesep}{1pt}
    \setlength{\belowrulesep}{1pt}
    \caption{
        Accuracy (\%) on cross-domain 5-shot node classification, reported as mean $\pm$ standard deviation. The best results are highlighted in bold, and the runner-up with an underline.
    }
    \label{tab:node_classification_5shot}

    \resizebox{\textwidth}{!}{%
    \begin{tabular}{@{}lcccccccc@{}}
        \toprule
        Method
        & Wisconsin & Cornell & CiteSeer & Cora
        & PubMed & Computers & Photo & Texas \\

        \midrule
        \multicolumn{9}{l}{\textit{Task-Supervised GNNs}} \\
        \midrule

        GCN
        & $27.00 \pm 6.60$ & $20.92 \pm 6.02$
        & $31.61 \pm 6.05$ & $39.48 \pm 6.08$
        & $38.82 \pm 2.98$ & $30.06 \pm 10.72$
        & $66.96 \pm 9.21$ & $38.08 \pm 13.65$ \\

        GAT
        & $28.85 \pm 8.91$ & $18.30 \pm 7.49$
        & $23.99 \pm 4.10$ & $29.82 \pm 2.31$
        & $34.92 \pm 12.27$ & $37.12 \pm 8.21$
        & $54.37 \pm 4.90$ & $28.21 \pm 11.78$ \\

        \midrule
        \multicolumn{9}{l}{\textit{Self-Supervised Pre-training with Fine-tuning}} \\
        \midrule
        DGI
        & $39.61 \pm 3.77$ & $38.92 \pm 6.78$
        & $53.66 \pm 6.21$ & $63.72 \pm 2.23$
        & $57.02 \pm 3.47$ & $52.03 \pm 5.29$
        & $69.74 \pm 4.95$ & $42.70 \pm 9.04$ \\

        GraphCL
        & $39.61 \pm 8.25$ & $42.70 \pm 5.54$
        & $39.26 \pm 3.63$ & $49.34 \pm 4.94$
        & $44.88 \pm 2.40$ & $57.78 \pm 3.96$
        & $69.39 \pm 6.46$ & $47.03 \pm 13.05$ \\

        \midrule
        \multicolumn{9}{l}{\textit{Prompt-based Adaptation}} \\
        \midrule
        GPPT
        & $35.29 \pm 7.72$ & $44.86 \pm 4.52$
        & $45.98 \pm 5.97$ & $48.78 \pm 5.49$
        & $48.88 \pm 8.25$ & $30.14 \pm 12.77$
        & $64.89 \pm 2.38$ & $48.11 \pm 10.71$ \\

        GPF
        & $33.73 \pm 3.37$ & $33.51 \pm 6.96$
        & $38.34 \pm 6.35$ & $49.98 \pm 5.98$
        & $43.00 \pm 3.48$ & $47.46 \pm 6.54$
        & $70.89 \pm 3.81$ & $30.81 \pm 12.86$ \\

        GraphPrompt
        & $31.37 \pm 3.10$ & $35.14 \pm 8.55$
        & $35.96 \pm 1.97$ & $48.24 \pm 3.43$
        & $58.56 \pm 2.40$ & $48.43 \pm 4.50$
        & $64.34 \pm 3.99$ & $27.03 \pm 3.82$ \\

        \midrule
        \multicolumn{9}{l}{\textit{Graph Foundation Models}} \\
        \midrule
        GCOPE
        & $34.38 \pm 3.93$ & $35.92 \pm 9.30$
        & $47.32 \pm 1.59$ & $57.33 \pm 2.08$
        & $60.17 \pm 1.81$ & $55.76 \pm 3.86$
        & $68.28 \pm 2.45$ & $26.34 \pm 10.63$ \\

        SAMGPT
        & $39.22 \pm 9.20$ & $41.08 \pm 11.13$
        & $59.44 \pm 5.70$ & $62.76 \pm 3.93$
        & $47.08 \pm 9.10$ & $58.81 \pm 7.14$
        & $72.08 \pm 4.20$ & $31.35 \pm 6.07$ \\

        MDGFM
        & $36.08 \pm 7.40$ & $41.08 \pm 7.72$
        & $57.08 \pm 3.71$ & $61.86 \pm 4.40$
        & $48.84 \pm 9.86$ & $49.68 \pm 4.93$
        & $70.37 \pm 7.36$ & $21.08 \pm 5.77$ \\

        RiemannGFM
        & $40.40 \pm 8.71$ & $44.86 \pm 4.05$
        & $54.68 \pm 3.67$ & $62.92 \pm 5.67$
        & $61.48 \pm 6.54$ & $48.93 \pm 5.75$
        & $73.47 \pm 6.42$ & $41.08 \pm 12.01$ \\

        R-GFM
        & $\underline{47.75 \pm 4.97}$ & $\underline{47.97 \pm 2.82}$
        & $\boldsymbol{74.59 \pm 1.87}$ & $\underline{63.77 \pm 2.47}$
        & $\underline{63.39 \pm 1.85}$ & $\underline{59.55 \pm 4.17}$
        & $\underline{73.62 \pm 1.15}$ & $\underline{51.64 \pm 6.31}$ \\

        \midrule
        \textbf{THETA}
        & $\boldsymbol{57.64 \pm 6.29}$ & $\boldsymbol{56.11 \pm 4.90}$ 
        & $\underline{64.97 \pm 3.34}$ & $\boldsymbol{66.96 \pm 3.81}$ 
        & $\boldsymbol{64.21 \pm 5.42}$ & $\boldsymbol{67.42 \pm 3.25}$ 
        & $\boldsymbol{78.23 \pm 3.98}$ & $\boldsymbol{55.48 \pm 5.84}$ \\
        \bottomrule
    \end{tabular}%
    }
\end{table*}

\subsection{Supervised Transferability (RQ2)}
\label{sec:link_prediction_results}

Table~\ref{tab:link_prediction} reports the supervised link prediction
results. THETA achieves the highest mean ROC-AUC on all four evaluated
datasets. Compared with the strongest reported baseline, R-GFM,
it improves ROC-AUC by 1.91, 4.75, 8.29, and 15.69 percentage points
on Wisconsin, Cora, PubMed, and Photo, respectively.
The gains cover Webpage, citation, and product co-purchase networks,
extending THETA's strong performance beyond node classification.

For link prediction, THETA incorporates HKS into the node
representations used for link scoring, together with node attributes
and encoder outputs. Heat diffusion thus provides a topology-based
structural prior that describes connectivity across diffusion scales
without relying on attribute semantics.
Unlike few-shot node classification, this task does not use prototype
refinement; the prior instead supplies structural context for predicting
node relationships. These results support the applicability of THETA's
heat-based design to supervised edge prediction across graph domains.

\begin{table}[t]
    \centering
    \renewcommand{\arraystretch}{0.88}
    \setlength{\tabcolsep}{3pt}
    \setlength{\aboverulesep}{1pt}
    \setlength{\belowrulesep}{1pt}
    \caption{
        Link prediction ROC-AUC
        (mean $\pm$ standard deviation, \%). The best results are highlighted in bold, and the runner-up with an underline.
    }
    \label{tab:link_prediction}

    \resizebox{\columnwidth}{!}{%
    \begin{tabular}{@{}lcccc@{}}
        \toprule
        Method & Wisconsin & Cora & PubMed & Photo \\
        \midrule

        GCOPE
        & $83.12 \pm 1.51$
        & $88.71 \pm 0.57$
        & $84.99 \pm 0.06$
        & $78.45 \pm 5.51$ \\

        SAMGPT
        & $65.32 \pm 7.27$
        & $88.62 \pm 3.51$
        & $84.92 \pm 0.40$
        & $80.59 \pm 0.33$ \\

        MDGFM
        & $67.46 \pm 6.82$
        & $88.41 \pm 2.12$
        & $84.78 \pm 2.09$
        & $81.24 \pm 5.80$ \\

        RiemannGFM
        & $78.19 \pm 2.08$
        & $88.54 \pm 0.11$
        & $86.37 \pm 0.33$
        & $81.44 \pm 0.95$ \\

        R-GFM
        & $\underline{84.15 \pm 0.71}$
        & $\underline{89.27 \pm 0.64}$
        & $\underline{88.62 \pm 0.41}$
        & $\underline{81.53 \pm 0.83}$ \\

        \midrule
        \textbf{THETA}
        & $\boldsymbol{86.06 \pm 3.14}$
        & $\boldsymbol{94.02 \pm 0.73}$
        & $\boldsymbol{96.91 \pm 0.25}$
        & $\boldsymbol{97.22 \pm 0.10}$ \\
        \bottomrule
    \end{tabular}%
    }
\end{table}

\subsection{Ablation Study (RQ3)}
\label{sec:rq3}

We conduct ablations on Cora, CiteSeer, and PubMed under
1-shot node classification. We separately remove graph
structure learning (GSL), upstream HKS, either the local
(shorter-time) or global (longer-time) HKS prior, and
downstream HKS. The scale-specific variants retain their
attribute branches, while the downstream variant retains
attribute-based adaptation and embedding-based retrieval.

\begin{figure}[tbp]
    \centering
    \includegraphics[width=\linewidth]
        {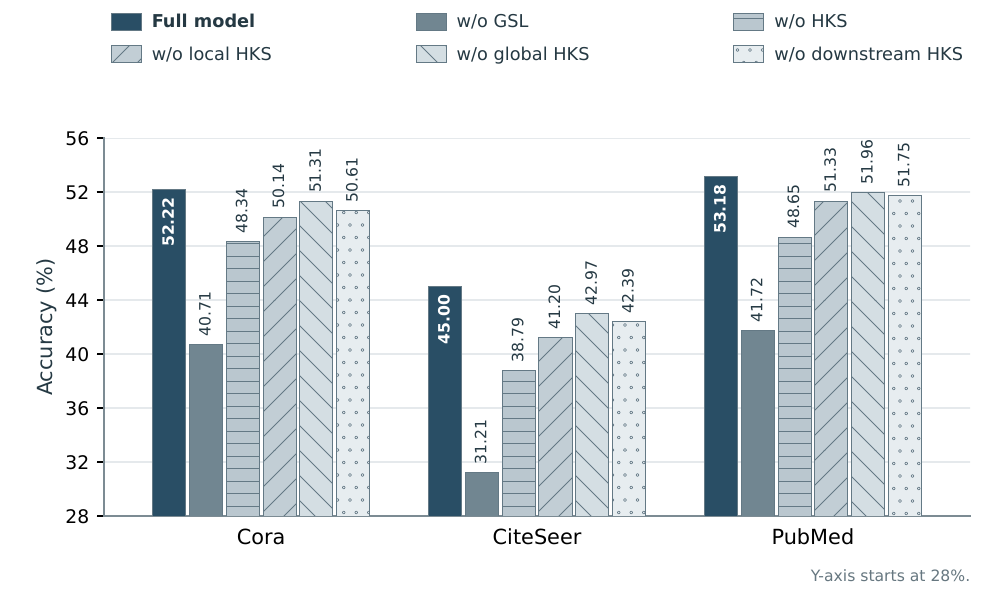}
    \caption{
        Ablation results for 1-shot node classification.
    }
    \label{fig:ablation}
\end{figure}

As shown in Figure~\ref{fig:ablation}, removing GSL causes
the largest accuracy decrease of 11.46--13.79 percentage
points, demonstrating the importance of learned graph
relations. Removing upstream HKS while retaining
attribute-based GSL reduces accuracy by 3.88--6.21 points.
Thus, heat diffusion provides useful structural guidance
beyond relations inferred from attributes, even when
HKS remains available during downstream adaptation.

Removing local HKS causes larger decreases than removing
global HKS on all three datasets, indicating a greater
predictive contribution from shorter diffusion scales
in this setting. Together with the cross-scale correlations
examined in Section~\ref{sec:rq4}, this suggests that the
two scale groups share substantial structural information,
while shorter-time descriptors provide more useful guidance
for these tasks. Nevertheless, removing global HKS also
consistently reduces accuracy, supporting its complementary
role rather than treating it as dispensable.

Removing downstream HKS reduces accuracy by 1.43--2.61
points despite retaining the full pre-trained model and
attribute-based adaptation. This shows that structural
guidance during pre-training does not fully replace the
value of explicitly using target-domain HKS.
Together, these results support heat diffusion as a
structural prior for both learning transferable relations
and adapting them with limited target labels.

\subsection{Cross-Scale Structure of the Heat Prior (RQ4)}
\label{sec:rq4}

\begin{figure}[t]
    \centering
    \includegraphics[width=\linewidth]
        {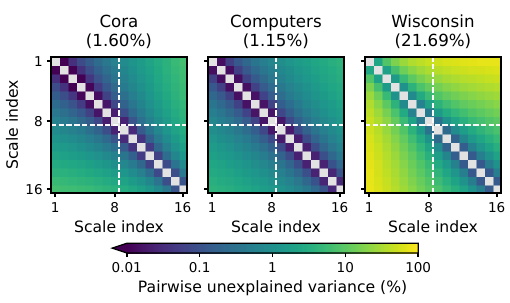}
    \caption{
        Cross-scale HKS differences with a shared logarithmic
        color scale.
    }
    \label{fig:hks_pairwise}
\end{figure}

We characterize the calibrated HKS prior using $100(1-r^2)$,
where $r$ is the Pearson correlation between two HKS scales
across nodes.
Figure~\ref{fig:hks_pairwise} shows that the median unexplained
variance between short- and long-time scales is only 1.60\%
on Cora and 1.15\% on Computers, but reaches 21.69\% on Wisconsin.
Thus, under the same calibration rule, cross-scale variation
remains graph-dependent: the two scale groups describe largely
shared structure on some graphs and more distinct variation
on others. This characterizes the topology-derived, multiscale
prior used by THETA; its contribution to prediction is assessed
separately in Section~\ref{sec:rq3}. Additional PCA analysis of cross-scale variation across all ten
graphs is provided in Figure~\ref{fig:hks_residual} in the appendix.

\subsection{Graph--Feature Balance in Target Adaptation (RQ5)}
\label{sec:rq5}

\begin{figure}[t]
    \centering
    \includegraphics[width=\linewidth]
        {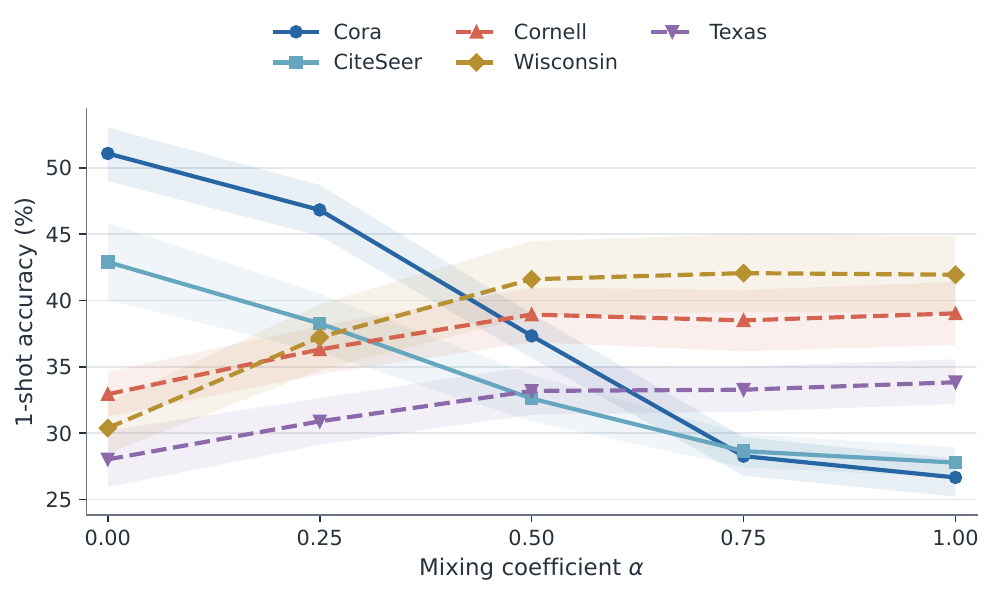}
    \caption{
        One-shot node classification accuracy versus the
        attribute-MLP mixing weight $\alpha$.
    }
    \label{fig:alpha_accuracy}
\end{figure}

We vary $\alpha$ in Eq.~\eqref{eq:target_node_embeddings}
under 1-shot node classification, independently adapting each
setting with matched support/query splits and initializations.
HKS-guided prototype refinement remains enabled.
Figure~\ref{fig:alpha_accuracy} shows that increasing the MLP
contribution benefits Cornell, Wisconsin, and Texas, with most
gains achieved by $\alpha=0.5$, but reduces accuracy on Cora
and CiteSeer.

\begin{figure}[t]
    \centering
    \includegraphics[width=\linewidth]
        {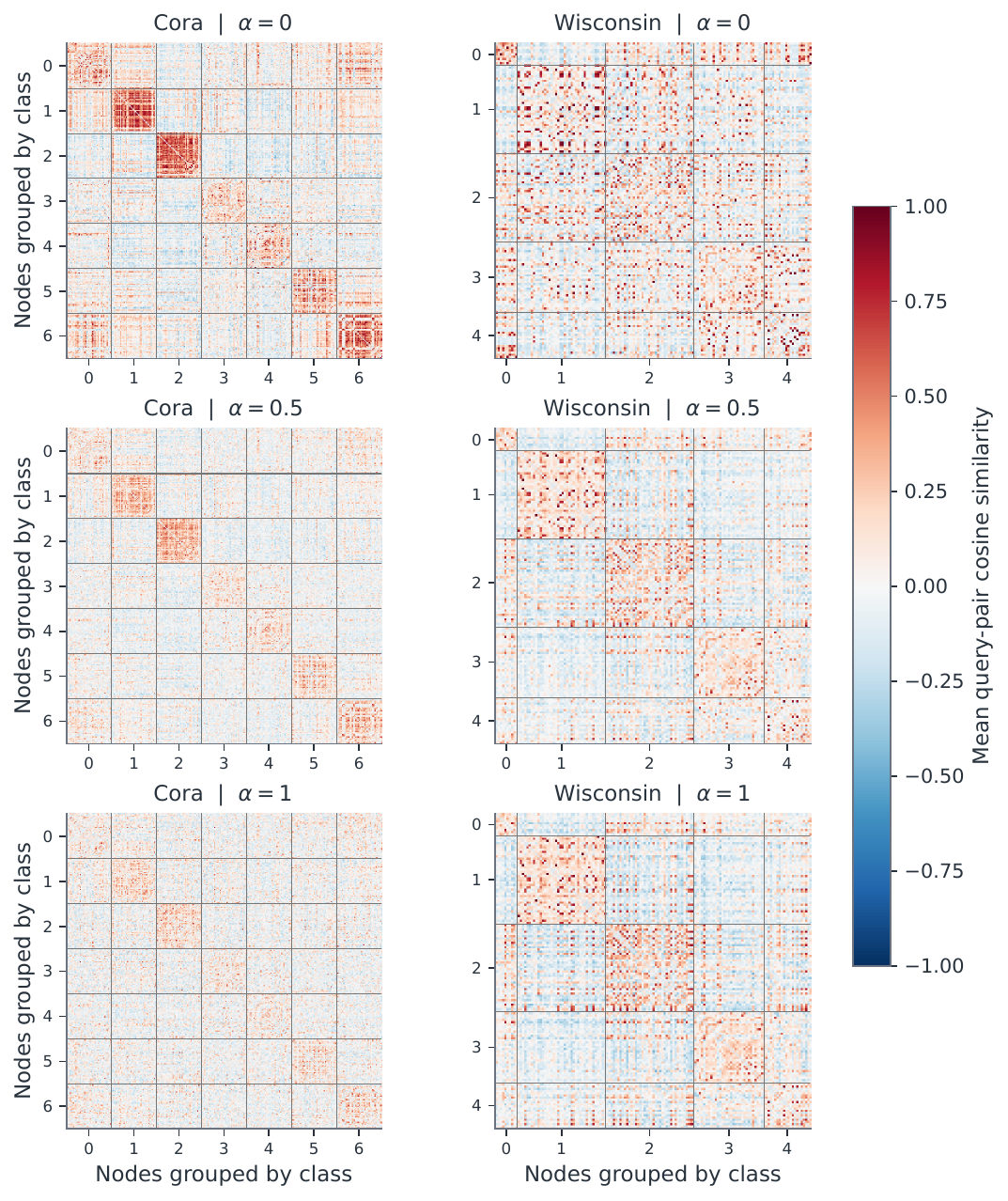}
    \caption{
        Class-ordered representation similarities on Cora
        (left) and Wisconsin (right) at $\alpha=0$, $0.5$,
        and $1$.
    }
    \label{fig:class_structure}
\end{figure}

The class-ordered similarities in Figure~\ref{fig:class_structure}
provide a complementary view: increasing $\alpha$ strengthens
within-class structure on Wisconsin but weakens it on Cora.
These patterns suggest that the heterophilous Webpage targets
rely more on direct attribute evidence, consistent with
neighborhood aggregation mixing features across class boundaries.
The MLP pathway can preserve discriminative attribute information
that such aggregation obscures. Conversely, the citation graphs
benefit more from propagation, consistent with informative
same-class neighborhoods.
Thus, effective adaptation combines the heat-based structural
prior with a target-dependent balance between propagation and
attribute evidence.
Further prediction-level analysis appears in
Figure~\ref{fig:error_decomposition} in the appendix.

\section{Conclusion}
\label{sec:conclusion}

We proposed THETA, a graph foundation model centered on heat
diffusion as an explicit structural prior for cross-domain
transfer. By describing multiscale heat-return behavior through
HKS, THETA derives structural evidence directly from graph
topology without requiring aligned attribute semantics.
This prior complements node attributes in learning graph
relations during pre-training and remains available to guide
edge-weight adjustment and prototype refinement during
few-shot adaptation. With the encoder frozen, THETA combines
target-domain structural information with learned representations
and limited support labels to adapt its propagation and use
of unlabeled nodes. Our work offers a perspective on GFMs
in which topology-derived knowledge serves as an explicit
reference throughout learning and transfer. It highlights
heat diffusion as a concrete way to connect structural
descriptions with the decisions underlying representation
learning and target-domain adaptation.

\bibliographystyle{ACM-Reference-Format}
\bibliography{ref}

\appendix

\section{Notation}
\label{app:notation}

Table~\ref{tab:notation} lists the main symbols, with domain
indices omitted when unambiguous.

\begin{table}[t]
    \centering
    \caption{Main notation.}
    \label{tab:notation}
    \small
    \setlength{\tabcolsep}{4pt}
    \begin{tabular}{@{}ll@{}}
        \toprule
        Symbol & Description \\
        \midrule
        $G_d,G_t$ & Source and target graphs \\
        $\mathbf{X}_d,\mathbf{A}_d$ & Attributes and adjacency \\
        $\mathcal{D}_{\mathrm{src}},\mathcal{G}_{\mathrm{src}}$ & Source domains and graphs \\
        $\mathbf{L}$ & Normalized Laplacian \\
        $\mathbf{K}_t$ & Heat kernel at time $t$ \\
        $h_v(t),\mathbf{h}_v$ & HKS value and node descriptor \\
        $\mathbf{H}^{(b)}$ & Branch-specific HKS matrix \\
        $M,M_s,M_l$ & Total, short-, long-time dimensions \\
        $b\in\{s,l\}$ & Short-/long-time branch \\
        $t_{G,j}$ & Calibrated diffusion time \\
        $m,r_G(t)$ & Nullity and normalized active trace \\
        $\rho_j,t_{\max}$ & Heat-trace target and time cap \\
        \midrule
        $f_\theta$ & Shared graph encoder \\
        $\mathcal{P}_d,\mathcal{P}_t$ & Source and target prompts \\
        $\mathbf{p}_{\mathrm{sh}},\mathbf{p}_d$ & Shared and source-domain tokens \\
        $\widetilde{\mathbf{X}}^{(b)},\widetilde{\mathbf{H}}^{(b)}$ & Prompted attributes and HKS \\
        $s_{uv}^{x,b},s_{uv}^{h,b}$ & Attribute and HKS similarities \\
        $g_{uv}^{(b)}$ & Relation fusion gate \\
        $\zeta$ & Learned-relation mixing weight \\
        $\mathbf{A}^{(o)},\mathbf{A}^{(b)}$ & Original and refined graph views \\
        $\mathcal{L}_{\mathrm{pre}}$ & Pre-training objective \\
        $\tau_h,\tau_{\mathrm{ctr}}$ & Kernel/contrastive temperatures \\
        \midrule
        $\overline{\mathbf{x}}_v,\overline{\mathbf{h}}_v$ & Fused target attributes and HKS \\
        $\delta_{uv}$ & Relative edge-weight adjustment \\
        $\mathbf{z}_v$ & Target node embedding \\
        $\alpha$ & Attribute-branch mixing weight \\
        $\mathcal{S}_c,\mathcal{U}$ & Support and unlabeled node sets \\
        $N,\mathcal{R}_c$ & Retrieval count and selected pairs \\
        $\lambda_{\mathrm{aug}},\tau_{\mathrm{aug}}$ & HKS retrieval weight/temperature \\
        $\mathbf{p}_c$ & Class prototype \\
        \bottomrule
    \end{tabular}
\end{table}

\section{Properties of the Heat-Diffusion Structural Prior}
\label{app:hks_properties}

This appendix gives calibration details, proves the fixed-time stability proposition in Section~\ref{sec:structural_prior}, and analyzes equivariance, time recalibration, and numerical approximation. The graph Laplacians in the exact analysis are real and symmetric, and all diffusion times are nonnegative. Matrix norms are spectral norms and vector norms are Euclidean norms.

\subsection{Diffusion-Time Calibration}
\label{app:hks_calibration}

For $n>m$, Eq.~\eqref{eq:normalized_heat_trace} can be written as an average over the positive Laplacian eigenvalues:
\begin{equation}
    \begin{aligned}
        r_G(t)&=\frac{1}{n-m}\sum_{\lambda_i>0}e^{-t\lambda_i},\\
        r_G'(t)&=-\frac{1}{n-m}\sum_{\lambda_i>0}
        \lambda_i e^{-t\lambda_i}<0.
    \end{aligned}
\end{equation}
Thus, $r_G(t)$ decreases continuously from $1$ toward $0$, and each target in $(0,1)$ determines a unique diffusion time before truncation. The decreasing targets in Eq.~\eqref{eq:heat_trace_targets} produce nondecreasing times. For targets reached within $t_{\max}$, the same coordinate corresponds to the same average decay level of the nonzero spectral modes across graphs, although the actual time is graph-specific. This calibration does not impose node correspondences or shared class meanings. If multiple targets require times beyond $t_{\max}$, they yield identical capped HKS coordinates, reducing the number of distinct diffusion scales.

If $n=m$, positive semidefiniteness implies $\mathbf{L}=\mathbf{0}$. There are no decaying modes, so the normalized heat trace in Eq.~\eqref{eq:normalized_heat_trace} is undefined. We set all diffusion times to zero and all HKS entries to one, matching the implementation. This convention also preserves equivariance, although the descriptors contain no variation across nodes or scales.

\subsection{Equivariance under Node Relabeling}
\label{app:hks_equivariance}

\textbf{Proposition (Equivariance under node relabeling).}
Let $\mathbf{H}(G)$ be the exact HKS matrix evaluated at the calibrated times $\mathcal{T}_G$. If $G^{\mathbf{P}}$ relabels $G$ by a permutation matrix $\mathbf{P}$, then
\begin{equation}
    \mathbf{H}(G^{\mathbf{P}})=\mathbf{P}\mathbf{H}(G).
    \label{eq:hks_permutation_equivariance}
\end{equation}
Thus, node relabeling only reorders the calibrated descriptors.

\emph{Proof.}
Let $G^{\mathbf{P}}$ have adjacency $\mathbf{A}'=\mathbf{P}\mathbf{A}\mathbf{P}^{\top}$. Its degree matrix is $\mathbf{D}'=\mathbf{P}\mathbf{D}\mathbf{P}^{\top}$, including the convention for zero-degree nodes. Thus
\begin{equation}
    \mathbf{L}'=\mathbf{P}\mathbf{L}\mathbf{P}^{\top},
    \qquad
    e^{-t\mathbf{L}'}=\mathbf{P}e^{-t\mathbf{L}}\mathbf{P}^{\top}.
\end{equation}
The number of nodes, nullity, and heat trace are unchanged, so $r_{G^{\mathbf{P}}}(t)=r_G(t)$. Using the same targets and time cap therefore gives $t_{G^{\mathbf{P}},j}=t_{G,j}$ for every $j$. Taking diagonals of the heat kernels at these times proves Eq.~\eqref{eq:hks_permutation_equivariance}.

\subsection{Stability at Fixed Diffusion Times}
\label{app:hks_fixed_stability}

We prove Eq.~\eqref{eq:hks_fixed_time_stability} from Section~\ref{sec:structural_prior}, using the same Laplacians, perturbation bound, and common time sequence.

\emph{Proof.}
For symmetric positive semidefinite $\mathbf{L}$ and $\widetilde{\mathbf{L}}$, their heat operators are contractions: $\|e^{-s\mathbf{L}}\|_2\leq1$ and $\|e^{-s\widetilde{\mathbf{L}}}\|_2\leq1$ for $s\geq0$. The matrix exponential identity gives
\begin{equation}
    \begin{aligned}
        &e^{-t\mathbf{L}}-e^{-t\widetilde{\mathbf{L}}}\\
        &\quad=\int_0^t e^{-(t-s)\mathbf{L}}
        (\widetilde{\mathbf{L}}-\mathbf{L})
        e^{-s\widetilde{\mathbf{L}}}\,ds.
    \end{aligned}
\end{equation}
Applying submultiplicativity and the contraction bounds yields
\begin{equation}
    \|e^{-t\mathbf{L}}-e^{-t\widetilde{\mathbf{L}}}\|_2
    \leq\int_0^t\epsilon\,ds=t\epsilon.
    \label{eq:heat_operator_stability}
\end{equation}
For the coordinate vector $\mathbf{e}_v$, the difference between the two diagonal entries is
\begin{equation}
    \begin{aligned}
        |h_v(t)-\widetilde{h}_v(t)|
        &=|\mathbf{e}_v^{\top}
        (e^{-t\mathbf{L}}-e^{-t\widetilde{\mathbf{L}}})
        \mathbf{e}_v|\\
        &\leq t\epsilon.
    \end{aligned}
\end{equation}
Squaring this inequality at each $t_j$, summing over $j$, and taking the square root proves Eq.~\eqref{eq:hks_fixed_time_stability}. This bound compares corresponding nodes under a Laplacian perturbation; it does not require a shared attribute space.

\subsection{Effect of Graph-Specific Diffusion Times}
\label{app:hks_clock_stability}

For symmetric normalized Laplacians, $\|\mathbf{L}\|_2\leq2$. Consequently, for any nonnegative $s,t$,
\begin{equation}
    \|e^{-t\mathbf{L}}-e^{-s\mathbf{L}}\|_2
    \leq2|t-s|,
    \label{eq:heat_time_stability}
\end{equation}
obtained by integrating the derivative $-\mathbf{L}e^{-u\mathbf{L}}$ between the two times. Let $\mathbf{t}_G=(t_{G,1},\ldots,t_{G,M})^{\top}$ and let $\mathbf{h}_{G,v}=\mathbf{h}_v(\mathbf{L};\mathcal{T}_G)$ denote the exact calibrated descriptor. Applying Eqs.~\eqref{eq:heat_operator_stability} and~\eqref{eq:heat_time_stability}, followed by the triangle inequality, gives
\begin{equation}
    \begin{aligned}
        \|\mathbf{h}_{G,v}-\mathbf{h}_{\widetilde{G},v}\|_2
        &\leq\epsilon\|\mathbf{t}_G\|_2\\
        &\quad+2\|\mathbf{t}_G-\mathbf{t}_{\widetilde{G}}\|_2.
    \end{aligned}
    \label{eq:hks_calibrated_stability}
\end{equation}
This separates changes in the heat operator from changes in the selected times. The second term requires control of the calibration procedure.

To make that dependence explicit, suppose $G$ and $\widetilde{G}$ have the same number of nodes and the same nullity $m<n$. Assume that, throughout $[0,t_{\max}]$, both normalized heat traces satisfy
\begin{equation}
    -r_G'(t)\geq\gamma,
    \qquad -r_{\widetilde{G}}'(t)\geq\gamma,
    \qquad \gamma>0.
    \label{eq:heat_clock_slope}
\end{equation}
Using $|\operatorname{tr}(\mathbf{B})|\leq n\|\mathbf{B}\|_2$ and Eq.~\eqref{eq:heat_operator_stability}, we obtain
\begin{equation}
    |r_G(t)-r_{\widetilde{G}}(t)|
    \leq\frac{nt\epsilon}{n-m}
    \leq\frac{nt_{\max}\epsilon}{n-m}.
\end{equation}
The slope condition bounds the sensitivity of the inverse heat traces, including their capped values. To see this, suppose $t_{G,j}<t_{\widetilde{G},j}$. The smaller time is uncapped and satisfies $r_G(t_{G,j})=\rho_j$. Since $r_{\widetilde{G}}(t_{\widetilde{G},j})\geq\rho_j$, Eq.~\eqref{eq:heat_clock_slope} gives
\begin{equation}
    \begin{aligned}
        \gamma(t_{\widetilde{G},j}-t_{G,j})
        &\leq r_{\widetilde{G}}(t_{G,j})-\rho_j\\
        &\leq\frac{nt_{\max}\epsilon}{n-m}.
    \end{aligned}
\end{equation}
The reverse ordering is handled by exchanging the graphs. Therefore,
\begin{equation}
    \|\mathbf{t}_G-\mathbf{t}_{\widetilde{G}}\|_2
    \leq\frac{\sqrt{M}\,nt_{\max}}{(n-m)\gamma}\epsilon.
    \label{eq:calibrated_time_bound}
\end{equation}
Together with $\|\mathbf{t}_G\|_2\leq\sqrt{M}\,t_{\max}$, this supplies a conditional bound for Eq.~\eqref{eq:hks_calibrated_stability}. A nearly flat heat trace can make $\gamma$ small and the bound loose. The argument also requires unchanged nullity, so it does not establish uniform stability under arbitrary changes in graph connectivity.

\subsection{Numerical Approximation}
\label{app:hks_numerical}

The preceding results concern exact heat operators and exact calibration. For large graphs, polynomial approximation and stochastic estimation affect both the selected times and the HKS values. Let $\widehat{\mathbf{t}}_G$ be the computed time vector, and let $\widehat{\mathbf{h}}_{G,v}$ be the computed descriptor. Write
\begin{equation}
    \eta_{G,v}
    =\|\widehat{\mathbf{h}}_{G,v}
    -\mathbf{h}_v(\mathbf{L};\widehat{\mathcal{T}}_G)\|_2
\end{equation}
for the error relative to the exact HKS at the computed times. Equation~\eqref{eq:heat_time_stability} then implies
\begin{equation}
    \|\widehat{\mathbf{h}}_{G,v}-\mathbf{h}_{G,v}\|_2
    \leq\eta_{G,v}
    +2\|\widehat{\mathbf{t}}_G-\mathbf{t}_G\|_2.
\end{equation}
Polynomial truncation and stochastic diagonal estimation both contribute to $\eta_{G,v}$; a truncation tolerance alone does not bound the stochastic error. Approximate permutation equivariance also depends on the probes: relabeling the probes together with the graph preserves the correspondence in exact arithmetic, whereas independently sampled probes introduce estimation differences. These numerical effects are separate from the exact structural properties and do not provide a guarantee of downstream prediction accuracy.


\section{Target-Domain Prompting}
\label{app:target_prompts}

We specify the target prompts used in
Section~\ref{sec:target_adaptation}. Each branch $b\in\{s,l\}$
has independent prompts for attributes and HKS. As in
Eq.~\eqref{eq:source_prompt}, let $\mathbf{q}$ denote an input
vector and temporarily omit branch and input-type indices.
The shared token $\mathbf{p}_{\mathrm{sh}}$ and source-domain
tokens $\{\mathbf{p}_d\}_{d\in\mathcal{D}_{\mathrm{src}}}$
are inherited from pre-training and remain frozen. We form
a source-derived modulation vector
\begin{equation}
    \mathbf{p}_{\mathrm{mix}}
    =\mathbf{p}_{\mathrm{sh}}\odot
    \left(\sum_{d\in\mathcal{D}_{\mathrm{src}}}
    a_d\mathbf{p}_d\right),
    \label{eq:target_source_token_mixture}
\end{equation}
where the coefficients $a_d\in\mathbb{R}$ are learned on
the target graph. A trainable target-specific token
$\mathbf{p}_t$ provides an additional modulation. The
target prompt is
\begin{equation}
    \begin{aligned}
        \mathcal{P}_t(\mathbf{q})
        &=\omega_t\mathbf{q}+(1-\omega_t)\\
        &\quad\cdot\bigl[
        (\beta_{\mathrm{src}}\mathbf{p}_{\mathrm{mix}}
        +\beta_{\mathrm{tgt}}\mathbf{p}_t)\odot\mathbf{q}\bigr],
    \end{aligned}
    \label{eq:target_prompt_mapping}
\end{equation}
where $\beta_{\mathrm{src}},\beta_{\mathrm{tgt}}\in\mathbb{R}$
are trainable fusion coefficients and
$\omega_t=\sigma(\kappa_t)\in(0,1)$ is controlled by a
trainable scalar $\kappa_t$. All tokens have the same
dimension as $\mathbf{q}$. The coefficients $a_d$ and
$\beta_{\mathrm{src}},\beta_{\mathrm{tgt}}$ are unconstrained;
they are not softmax-normalized or required to sum to one.
Disabling the residual connection corresponds to fixing
$\omega_t=0$. All target prompt parameters are separate
across branches and input types.

Let $\mathbf{x}_v^{\mathrm{PCA}}$ be the target node's
PCA-projected attributes. Restoring the branch and input-type
indices, the inputs used by downstream adaptation are
\begin{equation}
    \begin{aligned}
        \overline{\mathbf{x}}_v
        &=\frac{1}{2}\sum_{b\in\{s,l\}}
        \mathcal{P}_t^{(b,x)}(\mathbf{x}_v^{\mathrm{PCA}}),\\
        \overline{\mathbf{h}}_v
        &=\begin{bmatrix}
            \mathcal{P}_t^{(s,h)}(\mathbf{h}_v^{(s)})\\
            \mathcal{P}_t^{(l,h)}(\mathbf{h}_v^{(l)})
        \end{bmatrix}.
    \end{aligned}
    \label{eq:target_prompt_branch_fusion}
\end{equation}
Here, $x$ and $h$ identify attribute and HKS prompts,
respectively. Target prompting changes the modulation of
the structural coordinates, while the underlying HKS
descriptors remain fixed.

\section{Additional Experimental Details}
\label{app:experimental_details}

\subsection{Datasets}
\label{app:datasets}

Table~\ref{tab:dataset_statistics} summarizes the ten datasets.
Edges count distinct undirected pairs excluding self-loops;
Features denotes the attribute dimension before PCA.

\begin{table}[t]
    \centering
    \small
    \setlength{\tabcolsep}{3pt}
    \caption{Dataset statistics.}
    \label{tab:dataset_statistics}
    \begin{tabular}{@{}lrrrr@{}}
        \toprule
        Dataset & Nodes & Edges & Features & Classes \\
        \midrule
        Cora       & 2,708  & 5,278   & 1,433 & 7 \\
        CiteSeer   & 3,327  & 4,552   & 3,703 & 6 \\
        PubMed     & 19,717 & 44,324  & 500   & 3 \\
        \midrule
        Cornell    & 183    & 277     & 1,703 & 5 \\
        Texas      & 183    & 279     & 1,703 & 5 \\
        Wisconsin  & 251    & 450     & 1,703 & 5 \\
        \midrule
        Chameleon  & 2,277  & 31,371  & 2,325 & 5 \\
        Squirrel   & 5,201  & 198,353 & 2,089 & 5 \\
        \midrule
        Computers  & 13,752 & 245,861 & 767   & 10 \\
        Photo      & 7,650  & 119,081 & 745   & 8 \\
        \bottomrule
    \end{tabular}
\end{table}

\subsection{Baselines}
\label{app:baselines}

\paragraph{Supervised GNNs.}
GCN and GAT learn node representations through normalized
aggregation and attention, respectively.

\paragraph{Self-supervised methods.}
DGI and GraphCL learn representations without labels through
local--global information maximization and contrast between
augmented graph views, respectively.

\paragraph{Prompt-based methods.}
GPPT, GPF, and GraphPrompt adapt pre-trained graph models
through task-specific prompts.

\paragraph{Graph foundation models.}
GCOPE, SAMGPT, MDGFM, RiemannGFM, and R-GFM transfer knowledge
across domains using graph coordination, prompting, topology
learning, or Riemannian representations.

\subsection{Evaluation Protocols}
\label{app:evaluation_protocols}

We adopt leave-one-dataset-out transfer, pre-training on
the nine non-target graphs without node labels.
Node classification covers all datasets except Chameleon
and Squirrel; link prediction covers Wisconsin, Cora,
PubMed, and Photo.

\paragraph{Few-shot node classification.}
Each episode samples $K\in\{1,3,5\}$ support nodes per class
and uses the remaining labeled nodes as queries.
Target topology and attributes are available during adaptation,
while query labels are reserved for evaluation.
The encoder and source prompts remain frozen; target prompts,
the edge modifier, and the attribute MLP are optimized on
support labels.
HKS-guided prototype refinement is used for 1-shot evaluation,
whereas 3-shot and 5-shot evaluation use support-only prototypes.
Each episode reloads the pre-trained checkpoint and
reinitializes target modules.
We report the mean and standard deviation of final query accuracy.

\paragraph{Supervised link prediction.}
Positive edges are split into training, validation, and test
sets at an 80\%/10\%/10\% ratio, each paired with an equal
number of negatives sampled from nonedges of the original graph.
HKS is computed on the training observation graph.
We optimize the encoder, target prompts, and prediction head
with binary cross-entropy, keeping source prompts frozen.
The checkpoint maximizing validation ROC-AUC is evaluated
on the test set.
We report mean and standard deviation of test ROC-AUC
over repeated splits and initializations.

\subsection{Implementation Details}
\label{app:implementation}

\paragraph{Architecture and heat prior.}
We use a three-layer GCN with hidden and output dimensions
of 256 and graph-specific PCA with 50 output dimensions.
The HKS descriptor has 16 coordinates, equally divided
between short- and long-time branches.
Heat-trace targets decrease geometrically from 0.9 to 0.1,
with $t_{\max}=32$.
HKS computation uses eigendecomposition for graphs with
at most 4,096 nodes and stochastic polynomial approximation
for larger graphs.

\paragraph{Optimization and adaptation.}
We use Adam with a pre-training learning rate of
$3\times10^{-3}$ and weight decay of $10^{-5}$;
target adaptation uses a learning rate of $10^{-3}$.
The HKS kernel temperature, contrastive temperature, and
edge-dropout rate are 0.01, 0.2, and 0.3, respectively.
Prototype refinement retrieves up to $N=10$ candidates
per support node, with
$\lambda_{\mathrm{aug}}=\tau_{\mathrm{aug}}=0.5$.
Each support node receives twice the largest individual
candidate weight.

\begin{table}[t]
    \centering
    \small
    \setlength{\tabcolsep}{4pt}
    \caption{
        Reference implementation settings.
        Run-specific overrides should be reported separately.
    }
    \label{tab:implementation_settings}
    \begin{tabular}{@{}lr@{}}
        \toprule
        Hyperparameter & Value \\
        \midrule
        GCN layers & 3 \\
        Hidden/output dimension & 256 \\
        PCA dimension & 50 \\
        Short-/long-time HKS dimensions & 8 / 8 \\
        Heat-trace targets $(\rho_{\max},\rho_{\min})$
            & $(0.9,0.1)$ \\
        Maximum diffusion time $t_{\max}$ & 32 \\
        HKS kernel temperature $\tau_h$ & 0.01 \\
        Contrastive temperature $\tau_{\mathrm{ctr}}$ & 0.2 \\
        Edge-dropout rate & 0.3 \\
        Pre-training learning rate & $3\times10^{-3}$ \\
        Pre-training weight decay & $10^{-5}$ \\
        Adaptation learning rate & $10^{-3}$ \\
        Retrieval count $N$ & 10 \\
        Retrieval balance $\lambda_{\mathrm{aug}}$ & 0.5 \\
        Retrieval temperature $\tau_{\mathrm{aug}}$ & 0.5 \\
        \bottomrule
    \end{tabular}
\end{table}

\begin{figure}[tbp]
    \centering
    \includegraphics[width=\linewidth]
        {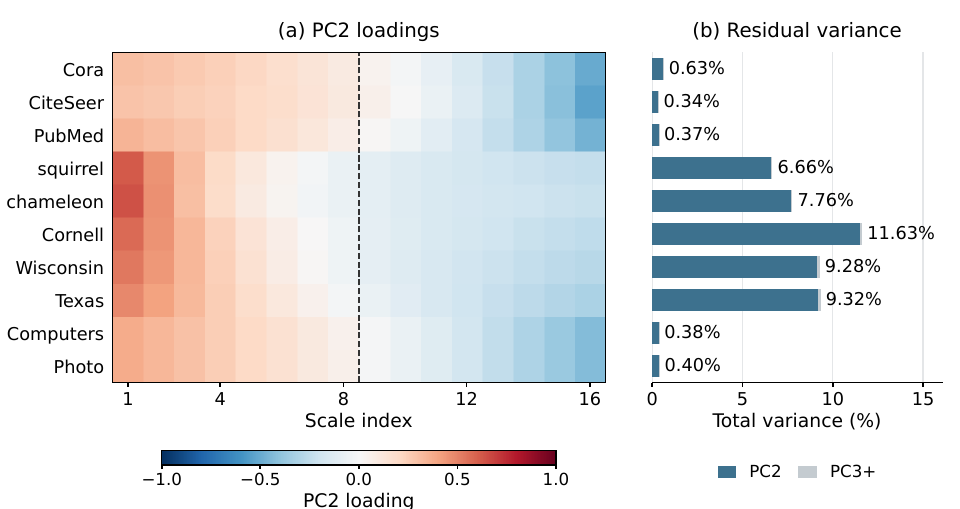}
    \caption{
        Residual cross-scale variation after per-scale
        standardization across nodes.
    }
    \label{fig:hks_residual}
\end{figure}

\begin{figure}[tbp]
    \centering
    \includegraphics[width=\linewidth]
        {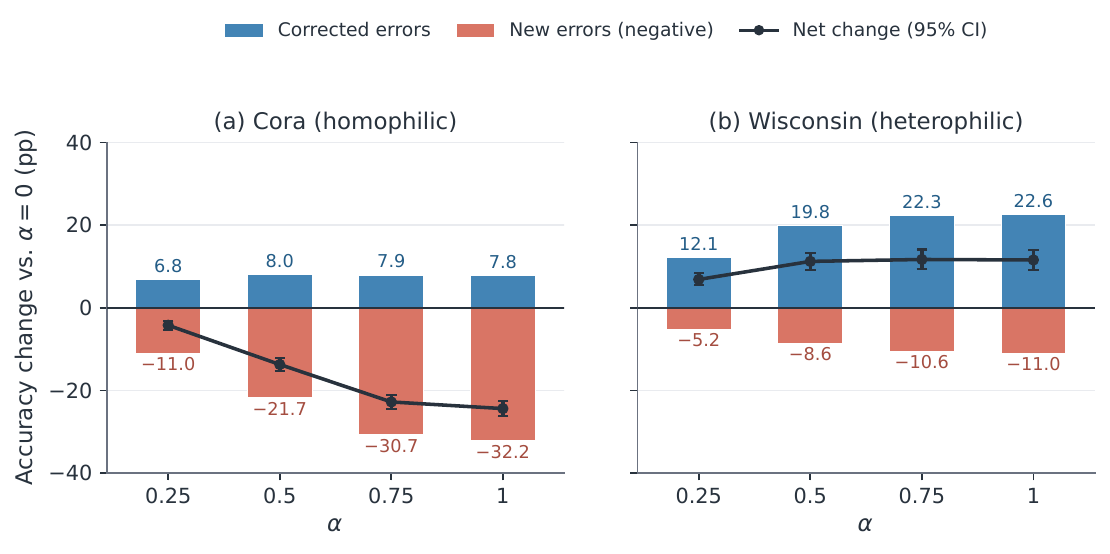}
    \caption{
        Prediction changes relative to $\alpha=0$ on Cora and
        Wisconsin.
    }
    \label{fig:error_decomposition}
\end{figure}

\end{document}